\documentclass[11pt]{amsart}
\usepackage{amsfonts,amsmath}

\newtheorem{definition}{Definition}

\usepackage{tikz}
\usetikzlibrary{shapes.geometric, arrows.meta, positioning, calc}

\tikzstyle{block} = [rectangle, minimum height=1.4cm, minimum width=3.8cm, text centered, draw=black, fill=blue!10, font=\sffamily\small, drop shadow]
\tikzstyle{arrow} = [thick,->,>=Stealth]
\tikzstyle{layer} = [rectangle, rounded corners, minimum height=1.2cm, minimum width=3cm, text centered, draw=black, fill=blue!10]
\tikzstyle{arrow} = [thick,->,>=Stealth]

\usepackage{enumerate,mathrsfs}
\usepackage{amssymb,amsmath,graphicx,amsthm,mathtools,hyperref}

\usepackage{booktabs}

\theoremstyle{plain}
\usepackage{siunitx}

\usepackage{algorithm}
\usepackage{algorithmic,colortbl}

\makeindex
\usepackage[intoc]{nomencl} 

\makenomenclature

\newcommand{\model}{\textsc{NeuroMemFPP}}

\usepackage{hyperref}
\hypersetup{nesting=true,debug=true,naturalnames=true}
\usepackage{graphicx,amssymb,upref}

\usepackage{subcaption}
\usepackage{enumerate,float}
\usepackage{fullpage}

\begin{document}

\title{\model: A \lowercase{Recurrent Neural Approach for Memory-Aware Parameter Estimation in Fractional} P\lowercase{oisson Process}}

\author{Neha Gupta}
\address{\emph{Operations Management and Quantitative Techniques Area, Indian Institute of Management Indore, Indore 453556, India.}}
\email{nehagupta@iimidr.ac.in}
\author[]{Aditya Maheshwari}
\address{\emph{Operations Management and Quantitative Techniques Area, Indian Institute of Management Indore, Indore 453556, India.}}
\email{adityam@iimidr.ac.in}

\email{}
\email{}

\begin{abstract}
In this paper, we propose a recurrent neural network (RNN)-based framework for estimating the parameters of the fractional Poisson process (FPP), which models event arrivals with memory and long-range dependence. The Long Short-Term Memory (LSTM) network estimates the key parameters $\mu >0$ and $\beta \in(0,1)$ from sequences of inter-arrival times, effectively modeling their temporal dependencies. Our experiments on synthetic data show that the proposed approach reduces the mean squared error (MSE) by about 55.3\% compared to the traditional method of moments (MOM) and performs reliably across different training conditions. We tested the method on two real-world high-frequency datasets: emergency call records from Montgomery County, PA, and AAPL stock trading data. The results show that the LSTM can effectively track daily patterns and parameter changes, indicating its effectiveness on real-world data with complex time dependencies.
\end{abstract}
\maketitle

 \section{Introduction and related work}
The classical or homogeneous Poisson process is a fundamental stochastic model for describing random event arrivals over time, with exponentially distributed inter-arrival times and a constant mean arrival rate \cite{Asmussen1987, Billingsley1979, Doob1953, Feller1971, Mikosch2009}. It has been widely applied in telecommunications \cite{chan1998outage}, queueing theory \cite{ross1978average, gross2011fundamentals}, neuroscience \cite{deger2012statistical, shinomoto2001modeling}, and finance \cite{ContTan2004, kluppelberg2004fractional}. However, many real-world systems exhibit long-range dependence, heavy-tailed waiting times, and non-Markovian temporal correlations, which the classical Poisson model cannot capture. To address these limitations, the fractional Poisson process (FPP) was introduced \cite{lask, mnv}, generalizing the classical model through fractional calculus. The FPP employs a Mittag–Leffler distribution for interarrival times, governed by a fractional parameter $\beta \in (0,1]$, allowing it to capture complex temporal patterns in real-world data.\\\\
The FPP has been applied in insurance, modeling claim arrivals more accurately than classical Poisson models \cite{kumar2019fractional}; in finance, capturing bursts and memory effects in high-frequency trading and transaction data \cite{kluppelberg2004fractional, ContTan2004}; and in environmental and sequential event data, such as wildfires, accounting for memory-dependent and subdiffusive behavior \cite{bapat2024modelling}. Its effectiveness depends on correctly estimating $\beta \in (0, 1)$ and $\mu \geq 0$, which control the frequency and clustering of events. Preliminary methods, including the method of moments (MOM) by Cahoy \cite{Cahoy2010}, addressed this problem, and recent studies show an effective fitting of the FPP to real-world data \cite{bapat2024modelling}.\\\\
Building on these developments, several recent works have explored data driven approaches for parameter estimation in stochastic models. In particular, neural networks have shown great promise: Convolutional Neural Network–Long Short-Term Memory (CNN–LSTM) architectures have been applied to L\'evy-driven stochastic differential equations \cite{lee2023recurrent}, recurrent networks to Hawkes processes on high-frequency financial data \cite{feng2023deep}, and deep learning methods to SDEs with fractional Brownian motion and measurement noise \cite{Wang2003}. Neural approaches have also been used for Stochastic Differential Equations (SDEs) with L\'evy noise \cite{wang2022neural} and for parameter estimation in Student L\'evy processes \cite{li2024parameter}.\\\\
Long Short-Term Memory (LSTM)-based Physics-Informed Neural Networks (PINNs) have recently been explored to improve parameter estimation by integrating sequential learning with physical constraints \cite{liu2023pi, ozalp2023physics, mahar2025attention}.  LSTM networks, a type of Recurrent Neural Network (RNN), are particularly well suited for sequential data \cite{hochreiter1997long, greff2016lstm}. Their ability to capture long-term dependencies makes LSTMs highly effective for modeling temporal patterns in stochastic processes \cite{sagheer2019time, pawar2018stock}, and particularly useful for capturing the memory-dependent dynamics of stochastic processes.\\\\
Inspired by recent advances in RNNs for parameter estimation in various fields \cite{wei2025estimating, wlas2008neural, wang2022neural} and the work of Lee et al. (2023) on Hawkes processes, we propose a direct RNN (LSTM)-based approach to estimate the parameters $\mu \geq 0$ and $\beta \in (0,1)$ of the FPP, which, to the best of our knowledge, has not been explored before, capturing its long-range dependency dynamics with promising results. In this paper, we make the following contributions:
\begin{itemize}
\item We propose an LSTM-based framework for direct parameter estimation of the FPP from sequential inter-arrival time data, which, to the best of our knowledge, has not been explored before.
\item Our experiments on synthetic data show that the LSTM-based method performs much better than the traditional method of moments, with 55.3\% lower MSE, six times faster speed, and greater stability under different training settings.
\item We validate our method on two real world datasets, emergency calls and financial transactions, and show that the LSTM model accurately tracks changing parameters and daily patterns in time based data with long term dependencies.
\end{itemize}
The rest of this paper is arranged as follows. Section \ref{section:back} provides a
background on the FPP and its distributional properties. Section \ref{section:3} describes our
LSTM-based methodology, including network architecture, data generation, and the
baseline MOM estimator. Section \ref{section:4} presents comprehensive experimental results on 
synthetic data, including convergence analysis, ablation studies, and computational 
efficiency comparisons. Section \ref{section:5} demonstrates applications to real emergency call 
and financial transaction datasets. Section \ref{section:6} concludes with a discussion of 
limitations and future research directions.

\section{Background}\label{section:back}
\subsection{Background}
Here, we define and present some preliminary results and definitions that will be used later in this paper.
\begin{definition}[Homogeneous Poisson process]
A L\'evy process $\{N(t),\; t \geq 0\},\; N(t) \in \mathbb{Z}^+$ is said to be homogeneous Poisson process or simply Poisson process with arrival rate $\mu >0$, if
\begin{enumerate}
\item $ \mathbb{P}(N(0) = 0) =1$.
\item The process has independent increments, that is, $N(s)$ is independent of $N(t+s)-N(s),  \; \forall \; s,t\geq 0$.
\item $\mathbb{P}(N(t+s)-N(s)=n) = e^{-\mu t}\frac{(\mu t)^n}{n!},\;n=0,1,\ldots,$ for all $s,\;t\geq 0.$
\item The paths of $N$ are $\mathbb{P}$-almost surely right continuous with left limits.
\end{enumerate}
\end{definition}
\begin{definition}[Fractional Poisson Process (FPP) {\cite{lask}}]
The Time-Fractional Poisson Process (FPP) was first introduced by Laskin (2003). It is defined via the following renewal process representation:
\begin{align}
N_{\beta}(t) = \max \left\{ n : T^{\beta}_{1} + T^{\beta}_{2} + \cdots + T^{\beta}_{n} \leq t \right\}, \quad t \geq 0,\; \beta \in (0,1],
\end{align}
where the inter-arrival times \( T^{\beta}_1, T^{\beta}_2, \ldots, T^{\beta}_n \) are i.i.d. non-negative random variables with the Mittag-Leffler distribution function:
\begin{align}
\mathbb{P}(T^{\beta}_k \leq x) = 1 - M_{\beta}(-\mu x^{\beta}), \quad x \geq 0,\; \mu > 0.
\end{align}
The probability density function (pdf) of each inter-arrival time \( T^{\beta}_k \) is given by:
\begin{align}
f(x) = \mu x^{\beta - 1} M_{\beta,\beta}(-\mu x^{\beta}), \quad x \geq 0,\; \mu > 0,\; \beta \in (0,1],
\end{align}
where $M_{a, b}(z)$ is the Mittag–Leffler function, defined as
\begin{equation}
    M_{a, b}(z) = \sum_{n=0}^{\infty} \frac{z^n}{\Gamma(an+b)}, \quad z \in \mathbb{C},
\end{equation}
where $a, b \in \mathbb{C}$ with $\Re(a) > 0$ and $\Re(b) > 0$
 
\end{definition}
\noindent\textbf{Distributional Properties of the FPP:} It is known that the mean and variance of the fractional Poisson process (FPP) are given by (see \cite{lask}):
\begin{align}
\mathbb{E}[N_\beta(t)] &= q t^\beta, \label{eq:mean_fpp} \\
\mathrm{Var}[N_\beta(t)] &= q t^\beta \left(1 + q t^\beta \left[ \beta B(\beta, 1/2) \cdot \frac{2}{2^{\beta} - 1} - 1 \right] \right), \label{eq:var_fpp1}
\end{align}
where \( q = \mu / \Gamma(1 + \beta) \), and \( B(a, b) \) denotes the Beta function.\\

\noindent\textbf{Long-memory and FPP:}
The concept and importance of long memory, also called long-range dependence (LRD), are well documented in the literature. Its relevance spans multiple application areas (see \cite{karag2004, Ding1993, Pagan1996, DouTaqqu2003, climate2006}). FPP was motivated by the observation of long-memory patterns in empirical data (see \cite{Cahoy2010}). The primary goal is to extend the classical Poisson model by allowing inter-arrival times to follow heavy-tailed, non-exponential distributions, thereby capturing more flexible scaling behaviours. This generalisation emphasises the temporal dependence between event arrivals in a counting process. Leonenko et al.\ (2014) (see \cite{leonenko2014correlation}) established the presence of the long-memory property in the fPP. Further, the studies in \cite{biardapp, lrd2016} demonstrated that the increments of the fPP exhibit long-range dependence characteristics.c

 \section{Methodology}\label{section:3}

In this section, we describe our LSTM-based framework for FPP parameter estimation. 
We first present the network architecture and explain how it processes sequential 
inter-arrival time data. We then describe our synthetic data generation procedure 
and introduce the baseline method of moments estimator used for comparison.
 \subsection{RNN (LSTM) Model Architecture}
We design an RNN based on the LSTM architecture to estimate the parameters of the FPP because the LSTM framework is particularly effective for modeling long-range dependence processes. It can capture long-term relationships in sequential data, while its gating structure enables the network to focus on the most relevant temporal patterns for accurate parameter estimation. Moreover, its recurrent design allows it to naturally handle sequences of varying lengths.\\
The objective of the model is to map sequences of inter-arrival times of the FPP to the underlying parameters $\mu > 0$ and $\beta \in (0, 1)$.
Each input is a sequence of shape $(L, 1)$, where:
\begin{itemize}
\item $L$ represents the sequence length (window size) used for generating the sample paths of the FPP, and
\item Each time step corresponds to the inter-arrival time between two consecutive events.
\end{itemize}

\begin{figure}[H]
    \centering    

    \includegraphics[width=1\linewidth]{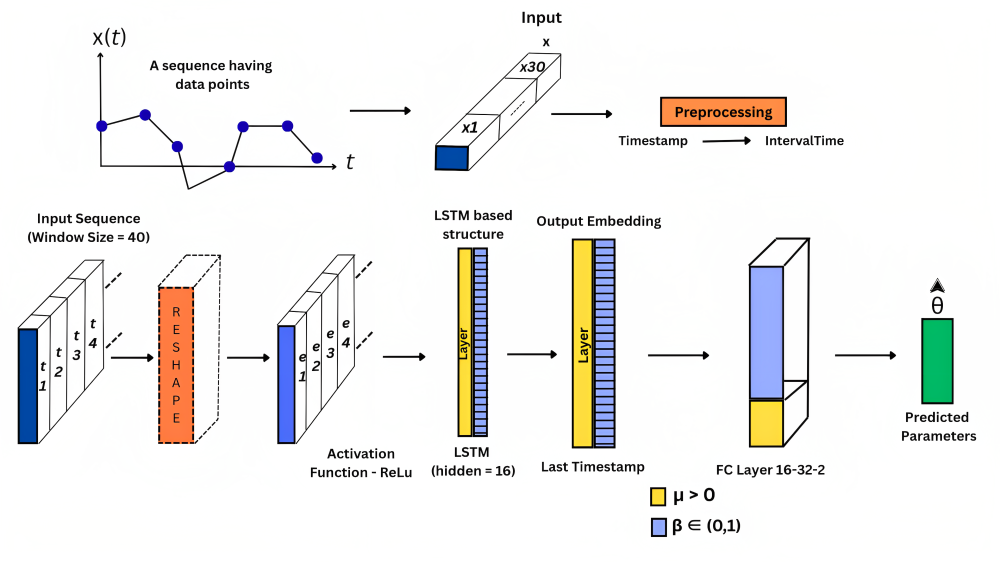}

    \begin{minipage}{.5\textwidth}
      \centering
      \includegraphics[width=.9\linewidth]{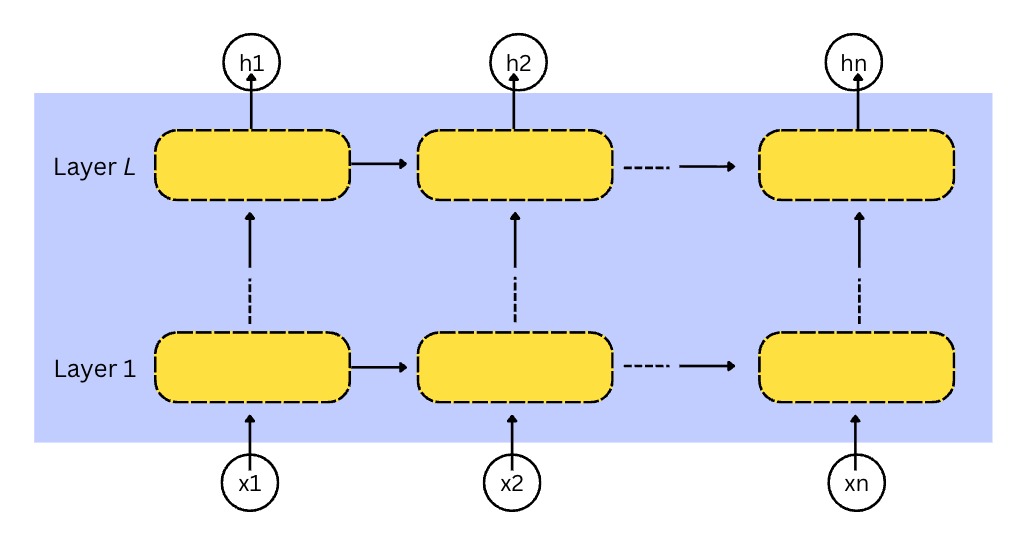}
    \end{minipage}%
    \begin{minipage}{.5\textwidth}
      \centering
      \includegraphics[width=.9\linewidth]{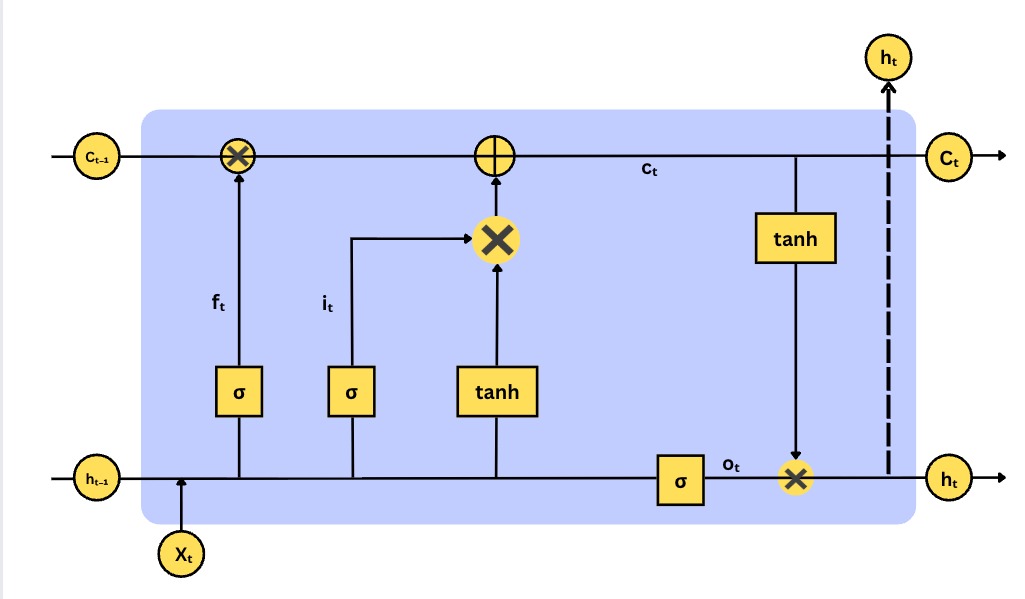}
    \end{minipage}
    \caption{LSTM-based architectures for parameter estimation 
    of the FPP: (top) main architecture for estimating 
    $\mu > 0$ and $\beta \in (0, 1)$; (bottom left/right) two 
    variants of the LSTM-based design.}
    \label{fig:all_arch}
\end{figure}

The model architecture is illustrated in Figure~\ref{fig:all_arch}. After preprocessing, timestamps are converted into inter-arrival times, which are windowed and reshaped into fixed-length input sequences. These sequences are passed through an activation function (ReLU) and then into an LSTM layer with 16 hidden units. The LSTM captures long-range temporal dependencies and nonlinear dynamics, making it suitable for modeling the memory-dependent structure of the FPP.

The LSTM layer computes its outputs using the following set of equations.
\begin{align*}
i_t &= \sigma(W_{ii} x_t + b_{ii} + W_{hi} h_{t-1} + b_{hi}) && \text{(input gate)} \\
f_t &= \sigma(W_{if} x_t + b_{if} + W_{hf} h_{t-1} + b_{hf}) && \text{(forget gate)} \\
g_t &= \tanh(W_{ig} x_t + b_{ig} + W_{hg} h_{t-1} + b_{hg}) && \text{(cell update)} \\
o_t &= \sigma(W_{io} x_t + b_{io} + W_{ho} h_{t-1} + b_{ho}) && \text{(output gate)} \\
c_t &= f_t \odot c_{t-1} + i_t \odot g_t && \text{(cell state update)} \\
h_t &= o_t \odot \tanh(c_t) && \text{(hidden state output)}
\end{align*}

Here, $i_t$, $f_t$, $g_t$, and $o_t$ represent the input, forget, cell, and output gates, respectively, at time $t$. The cell state $c_t$ and hidden state $h_t$ capture the temporal memory of the sequence. The symbol $\odot$ denotes element-wise multiplication, and $\sigma$ represents the \texttt{sigmoid} activation function.
At the end of the sequence, the final hidden state $h_t$ serves as the feature representation summarizing the entire input. This is passed through a fully connected (FC) network with layer dimensions 16–32–2, using \texttt{ReLU} activation between layers. To ensure the predicted parameters remain in valid ranges, the \texttt{softplus} activation is applied for $\mu > 0$ and the \texttt{sigmoid} activation for $\beta \in (0, 1)$. The final output is the predicted parameter vector $(\hat{\mu}, \hat{\beta})$ for each input sequences.

\noindent The traditional method for estimating the parameters $\beta$ and $\mu$ of the FPP is the method of moments (MOM) \cite{Cahoy2010}, which matches theoretical moments with the corresponding sample moments derived from the data. 
\subsection*{Data generation}
For this study, we generated synthetic data to evaluate the performance of our RNN/LSTM-based parameter estimation model. The data were produced by simulating sample paths of the Fractional Poisson Process (FPP) using a transformation-based algorithm. For each experiment, specific values of the parameters $\mu$ and $\beta$ were chosen, and for each pair, $n$ inter-arrival times were generated using three independent random variables $U_1$, $U_2$, and $U_3$ sampled from a uniform distribution between 0 and 1. The algorithm computes an auxiliary variable

\[
S = \frac{\sin(\beta \pi U_2) \, [\sin((1 - \beta)\pi U_2)]^{1/\beta - 1}}{[\sin(\pi U_2)]^{1/\beta} \, |\log(U_3)|^{1/(\beta - 1)}}
\]

and then the inter-arrival time

\[
T = \frac{|\log(U_1)|^{1/\beta}}{\mu^{1/\beta}} \times S.
\]

Repeating this procedure $n$ times yields a sequence of inter-arrival times $\{T_1, T_2, \dots, T_n\}$, whose cumulative sums define the event times of an FPP sample path. These cumulative times capture the stochastic behavior of fractional waiting times governed by $\mu$ and $\beta$. The generated sequences of inter-arrival times serve as inputs to the neural network, while the corresponding parameter values $(\mu, \beta)$ are used as target labels for supervised training.

To evaluate the performance of our RNN (LSTM) in estimating parameters, we compare the parameters predicted by the model with the classical MOM estimators \cite{Cahoy2010}. The MOM approach estimates the parameters most likely to generate the observed data by equating theoretical moments with the sample moments derived from the inter-arrival times $T$. The estimators obtained are given by:

\begin{align*}
\hat{\beta} &= \frac{\pi}{r^3} \left( \sigma_d^2 \ln T + \frac{\pi^2}{6} \right)\\
\hat{\mu} &= \exp \left( -\hat{\beta} \left( \mathbb{E}[\ln(T)] + C \right) \right),
\end{align*}

where $\hat{\beta}$ and $\hat{\mu}$ represent the MOM-based estimates of the parameters $\beta$ and $\mu$, respectively.

 \section{Experimental setup}\label{section:4}
 
\begin{figure}[H]
    \centering
    \includegraphics[width=0.7\linewidth]{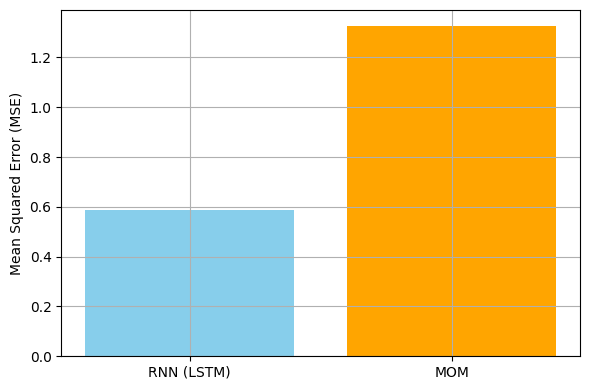}
    \caption{Comparison of parameter estimation accuracy for the FPP using RNN and MOM. The RNN model significantly performance well the method of moments, achieving a much lower Mean Squared Error}
    \label{RMSE_RNN_TFPP}
\end{figure}

 In this study, we conducted a series of experiments to evaluate the LSTM's performance on synthetic 
data. Our experimental setup, comparison with the baseline MOM estimator, convergence analysis, ablation studies, and computational efficiency 
assessment.
The synthetic dataset was created by simulating inter-arrival times from the FPP for a wide range of parameter values. For each simulation, the parameters $\mu >0$ and $\beta \in(0,1)$ were randomly drawn from uniform distributions within the ranges 
$[0.5,5.0]$ and 
$[0.1,0.9]$, respectively. Each generated sequence consisted of 50 inter-arrival events, and a total of 100,000 such samples were produced. The cumulative event times were computed to form temporal sequences, while their corresponding inter-arrival increments were used as the primary input features for training. Therefore, each sample contained both the event-time sequence and its true underlying parameters $(\mu, \beta)$, which served as targets for supervised learning.\\
The proposed RNN (LSTM) architecture was designed to capture the sequential dependencies in the FPP data. The model consisted of a single LSTM layer with $16$ hidden units, followed by a fully connected layer with $32$ neurons using the \texttt{ReLU} activation function, and an output layer with two neurons corresponding to the predicted parameters $\mu$ and $\beta$. To ensure valid parameter predictions, the output activations were chosen such that
$\mu >0$ (a \texttt{softplus} function), $\beta \in (0,1)$ (a \texttt{sigmoid} function). The data set was divided into 80\% for training and 20\% for testing. Training was performed for 100 epochs using the Adam optimizer \cite{kingma2014adam} with a learning rate of $0.001$, and the mean squared error (MSE) was used as the loss function.\\
After training, the model was evaluated on the test dataset, and its mean squared error (MSE) performance was compared with that of the MOM estimator. In the MOM approach, a numerically stable implementation was employed to prevent computational errors by clipping extreme values and handling unstable logarithmic transformations. Only valid parameter estimates were retained for evaluation. The comparison involved computing the MSE between the true parameters and those predicted by both methods. Based on the results shown in Figure~\ref{RMSE_RNN_TFPP}, the RNN (LSTM) method reduces the MSE by approximately 55.3\% compared to the MoM. This demonstrates that the RNN (LSTM) approach provides significantly higher accuracy in estimating the parameters of the FPP, emphasizing the effectiveness of neural network–based estimation techniques.\\
Furthermore, we analyzed several results to demonstrate the performance of the RNN (LSTM) model for parameter estimation of the FPP. These results include the variation of the root mean square error (RMSE), the mean absolute error (MAE), and the coefficient of determination ($R^2$) with respect to the number of training epochs, sample sizes, sequence lengths, learning rates, batch sizes and hidden sizes.\\

\begin{figure}[H]
\centering
\begin{minipage}{.45\textwidth}
\centering
\includegraphics[width=1\linewidth]{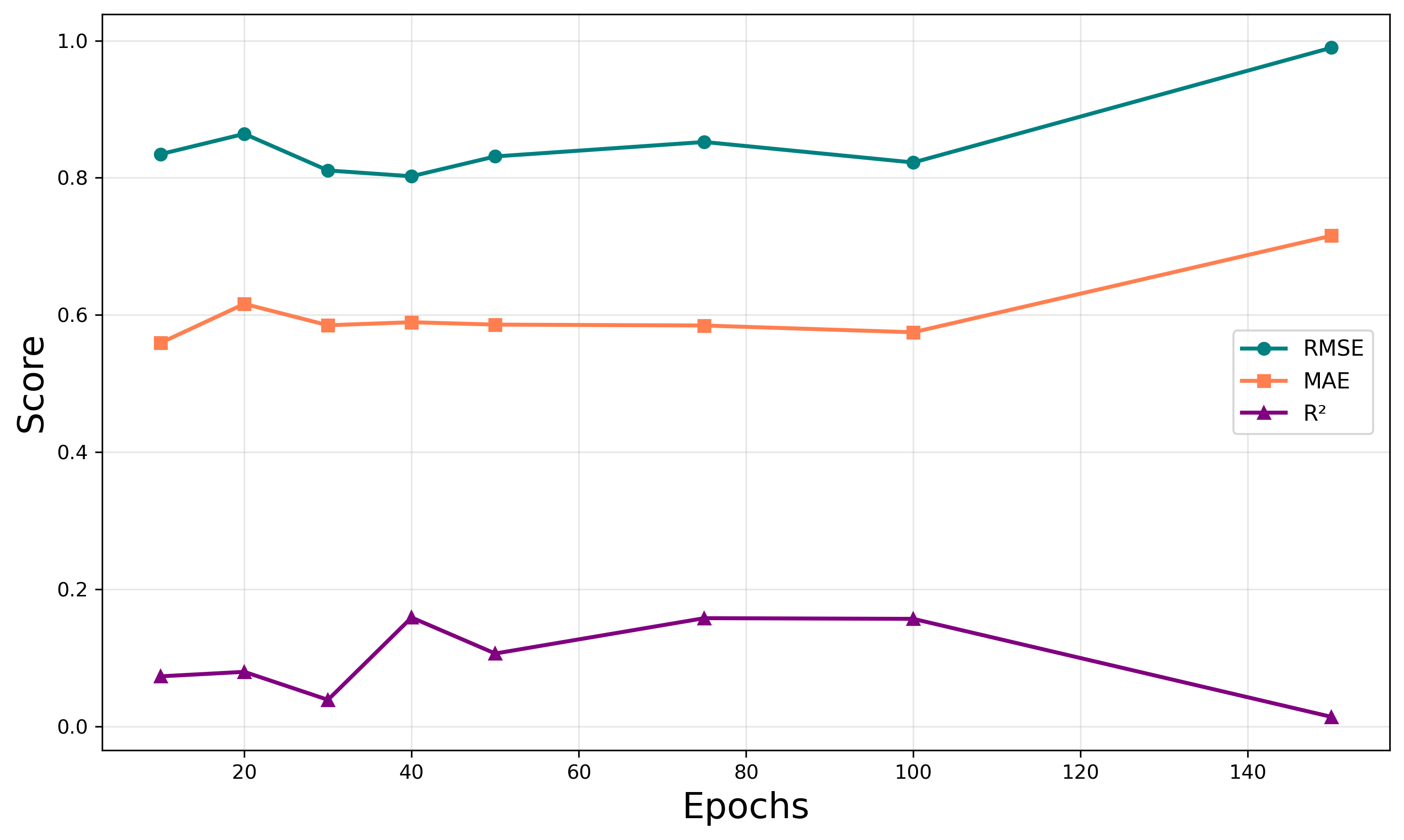}
\end{minipage}%
\begin{minipage}{.45\textwidth}
\centering
\includegraphics[width=1\linewidth]{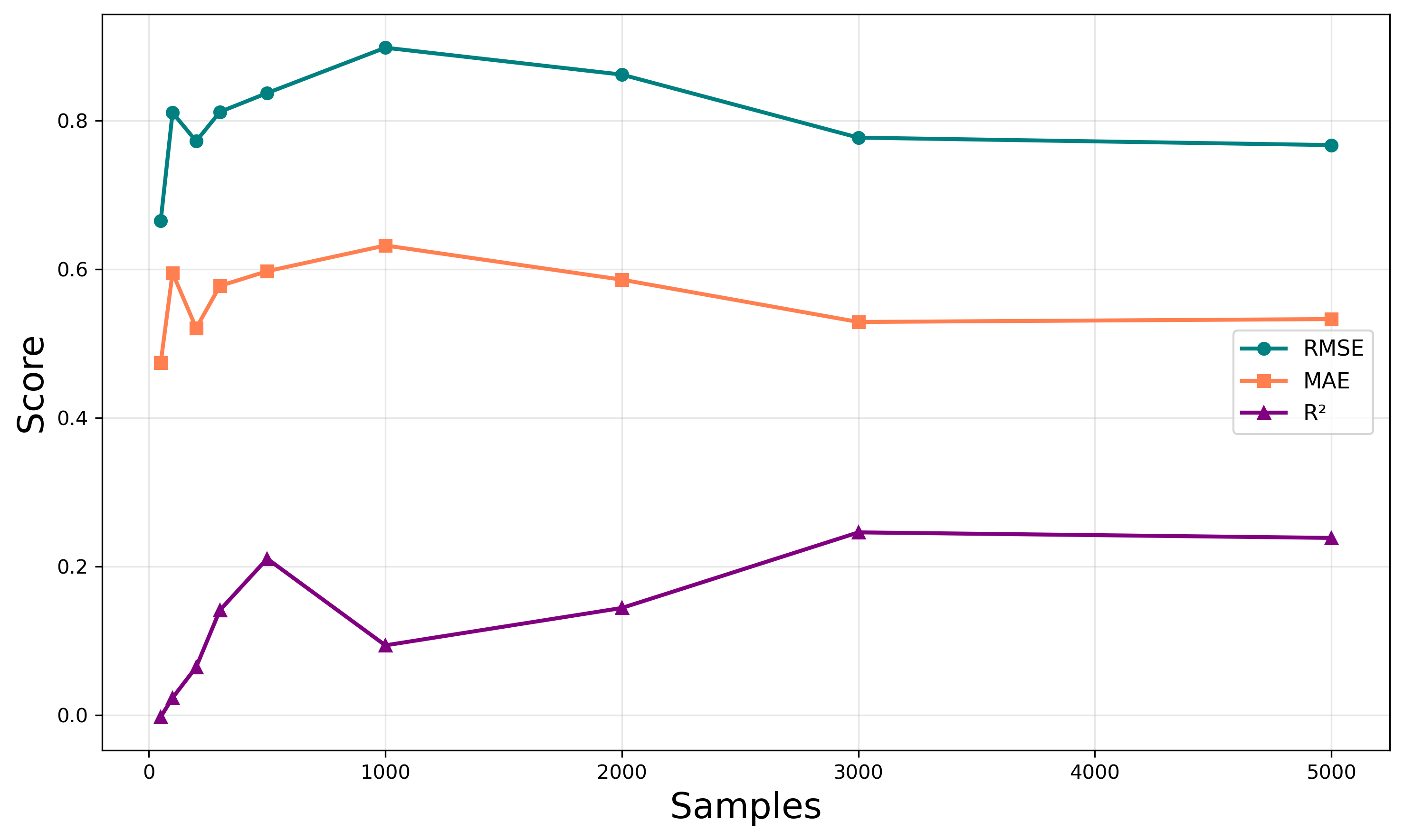}
\end{minipage}
\caption{Effect of training epochs (left) and sample sizes (right) on the RMSE, MAE, and $R^2$ of RNN (LSTM)-based FPP parameter estimation.}
\label{epoches_Samples}
\end{figure}

The figure on the right in Figure~\ref{epoches_Samples} presents the relationship between the number of training samples and the evaluation metrics (RMSE, MAE, and $R^2$). The results indicate that with very small sample sizes, the model exhibits poor performance, with high error values (RMSE $\approx 1.0$, MAE $\approx 0.67$) and a negative $R^2$ value ($\approx -0.4$). As the number of samples increases, both RMSE and MAE gradually decrease, while $R^2$ improves and becomes positive around 500 samples. Beyond 2000 samples, the metrics stabilize, with RMSE $\approx 0.78$, MAE $\approx 0.53$, and $R^2 \approx 0.23$. This suggests that although larger datasets help reduce prediction error, the explained variance remains relatively moderate.\\

The figure on the left in Figure~\ref{epoches_Samples} illustrates the effect of training epochs on RMSE, MAE, and $R^2$. Initially, RMSE is around $1.0$ and stabilizes near $0.92$, while MAE fluctuates between $0.55$ and $0.70$ before converging close to $0.60$. The $R^2$ score improves slightly during the initial epochs (reaching about $0.12$ at epoch $50$) but remains unstable and eventually returns close to zero by epoch $200$. These results indicate that while the model benefits significantly from an increase in training data, extending the number of epochs beyond a certain point yields limited performance gains.

\begin{figure}[H]
\centering
\begin{minipage}{.45\textwidth}
\centering
\includegraphics[width=1\linewidth]{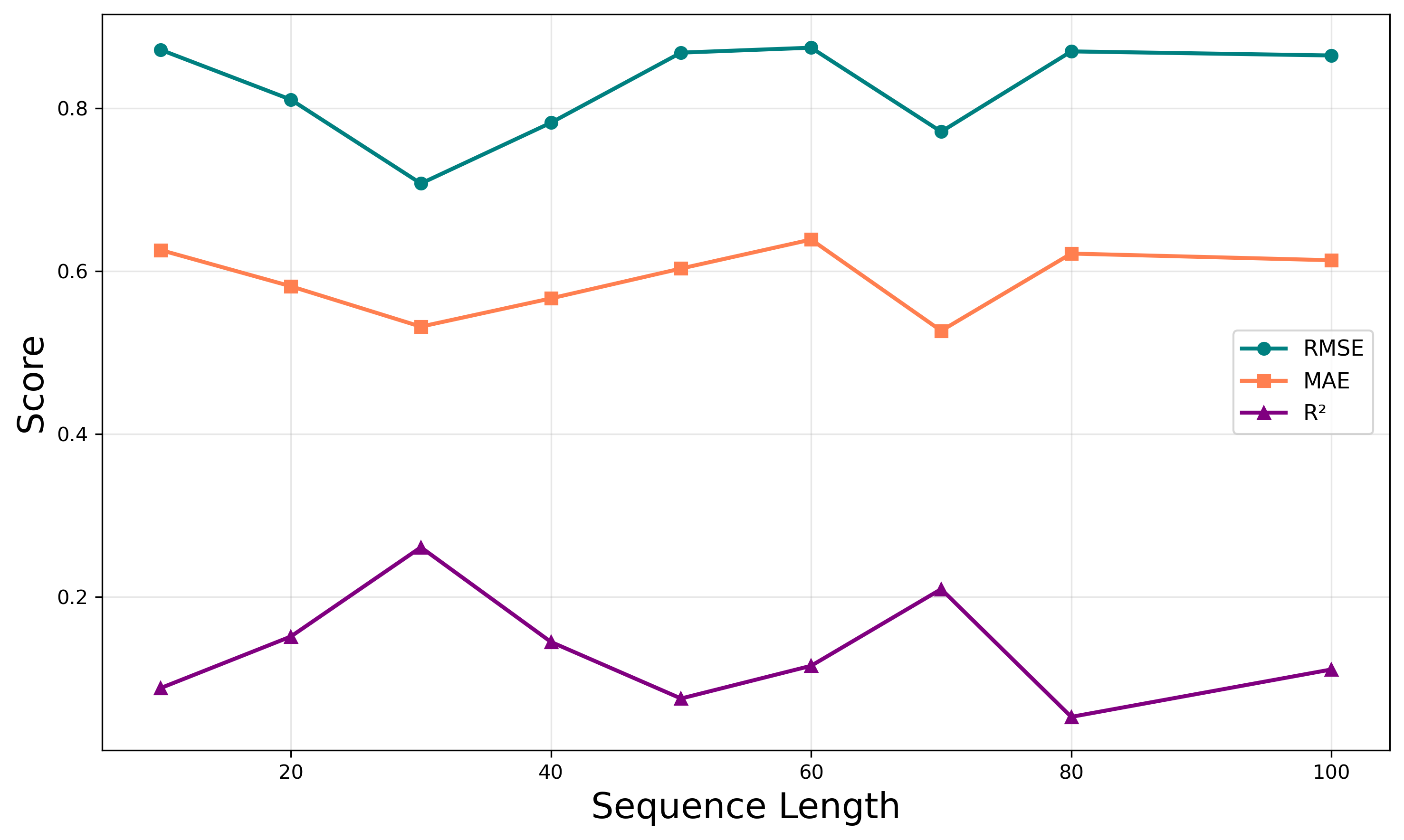}
\end{minipage}%
\begin{minipage}{.45\textwidth}
\centering
\includegraphics[width=1\linewidth]{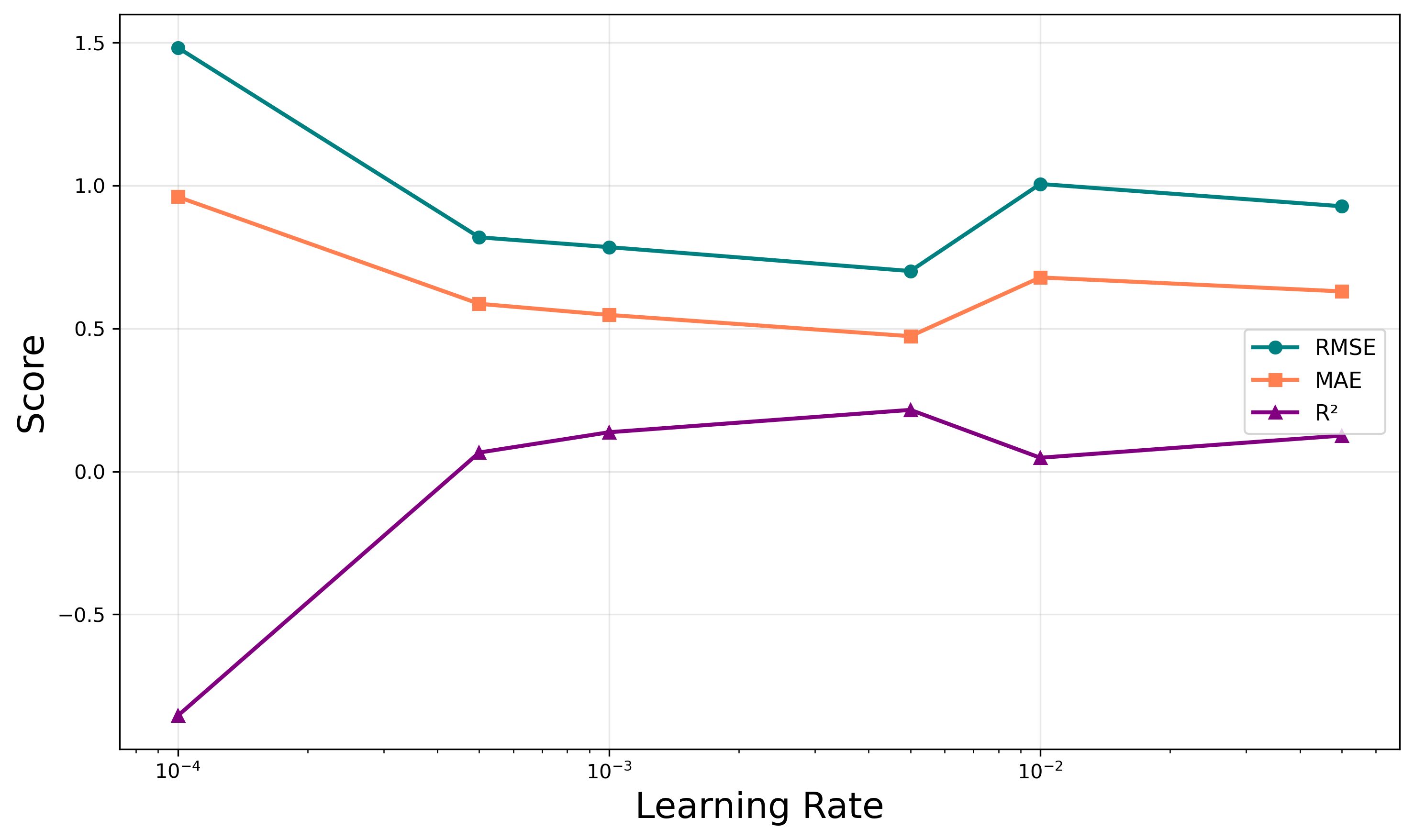}
\end{minipage}
\caption{Effect of sequence length (left) and learning rate (right) on the RMSE, MAE, and $R^2$ of RNN (LSTM)-based FPP parameter estimation.}
\label{seq_learning}
\end{figure}

Next, Figure~\ref{seq_learning} presents the combined effect of sequence length (left) and learning rate (right) on model performance. Increasing the length of the sequence generally improves accuracy, with the RMSE decreasing from 0.89 to 0.76 and the MAE from 0.63 to 0.56, while the $R^{2}$ score peaks around 0.27, indicating a better temporal representation. The learning rate analysis shows that mid-range values ($10^{-3}$ to $5\times10^{-3}$)  yield the best results, achieving lower RMSE ($\approx 0.76$-$0.88$), lower MAE ($\approx 0.52$-$0.63$), and stable positive $R^{2}$ values. Very low or very high learning rates degrade performance, demonstrating the importance of moderate learning rate.

\begin{figure}[H]
\centering
\begin{minipage}{.45\textwidth}
\centering
\includegraphics[width=1\linewidth]{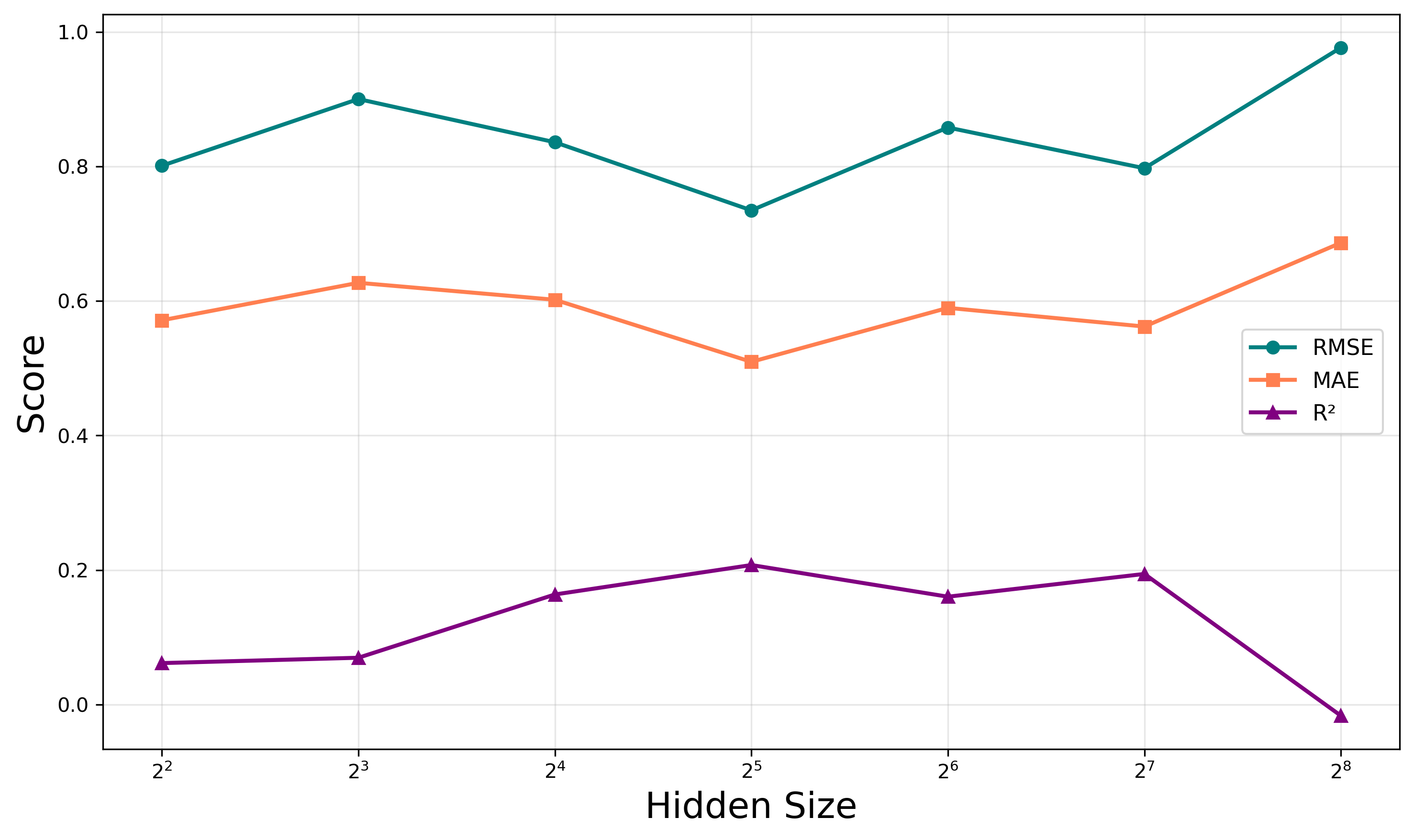}
\end{minipage}%
\begin{minipage}{.45\textwidth}
\centering
\includegraphics[width=1\linewidth]{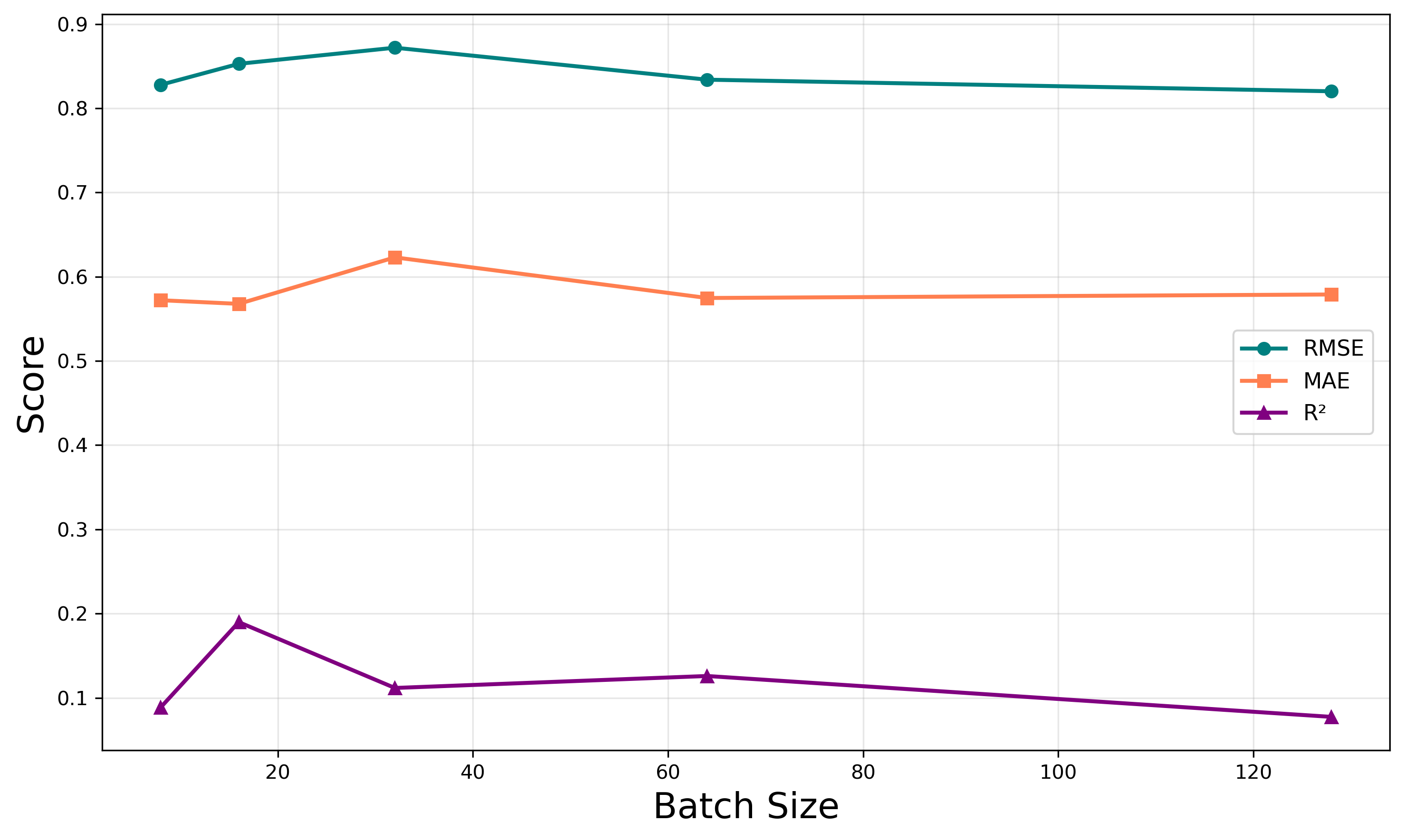}
\end{minipage}
\caption{Effect of hidden size (left) and batch size (right) on the RMSE, MAE, and $R^2$ of RNN (LSTM)-based FPP parameter estimation.}
\label{hidden_batch}
\end{figure}

Moreover, Figure~\ref{hidden_batch} show how hidden size and batch size influence the model's performance. When the hidden size increases from $2^{2}$ to $2^8$, the performance changes only slightly, RMSE remains within $0.76-0.97$, MAE within $0.50-0.70$, and the $R^2$ score stays low (from $0.02$ to $0.17$). In contrast, increasing the batch size from $8$ to $128$ leads to only a slight performance change, RMSE shifts from approximately $0.83$ to $0.82$, MAE remains almost constant around $0.57$–$0.58$, and the value of $R^2$ decreases only slightly from $0.09$ to $0.08$. This indicates that hidden size and batch size have only a minor impact on the model’s performance.

\begin{figure}[H]
\centering
\includegraphics[width=0.7\linewidth]{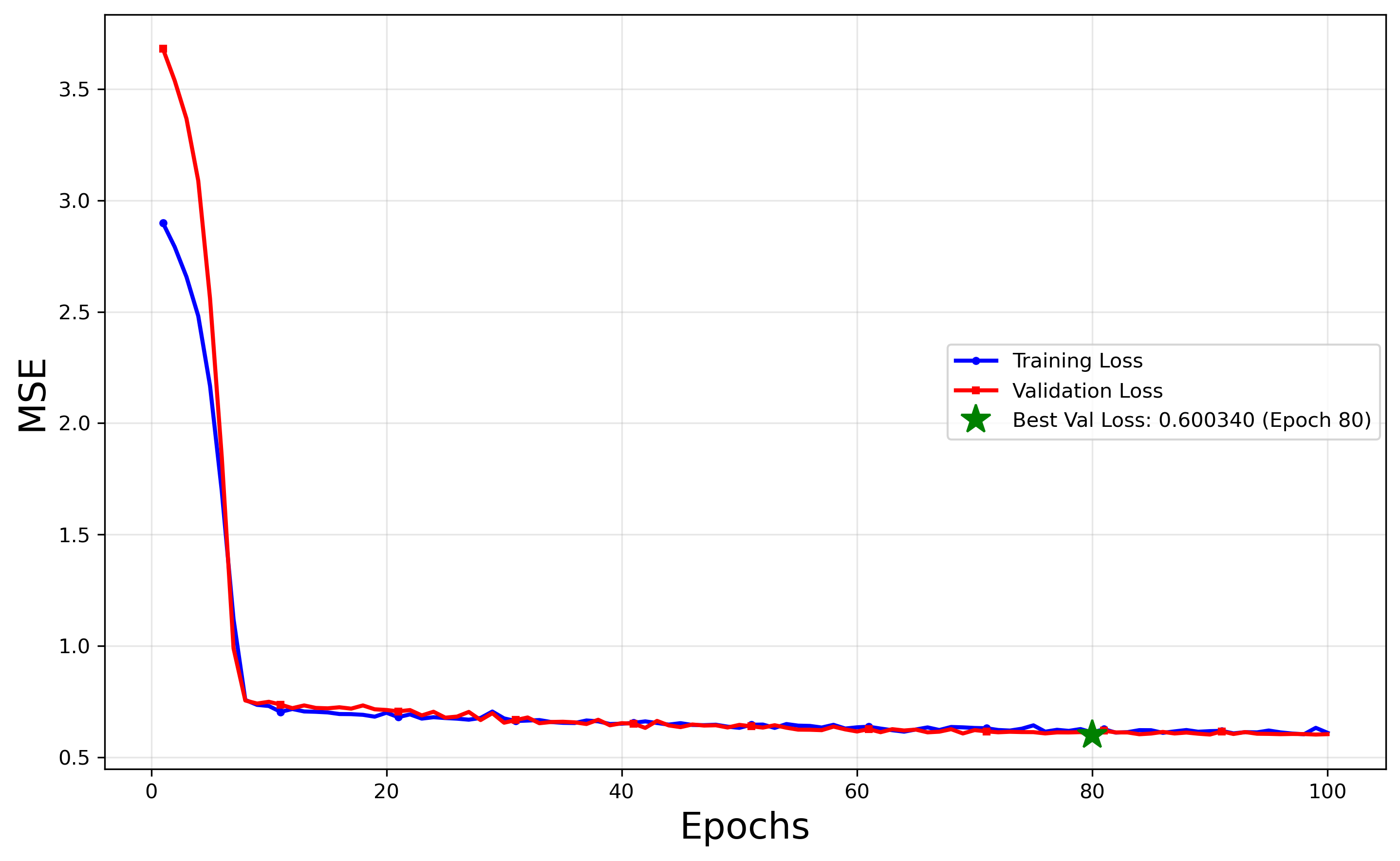}
\caption{Convergence of training and validation loss, showing minimum validation loss at epoch 80.}
\label{RMSC_Epochs}
\end{figure}

Additionally, the training and validation losses were analyzed as functions of epochs to study the model’s convergence behavior, as shown in Figure~\ref{RMSC_Epochs}. Both training and validation losses decrease rapidly during the initial epochs and then stabilize around $0.65$, indicating that the model learns efficiently in the early stages and achieves consistent convergence. The lowest validation loss is observed around epoch $80$, suggesting this as the optimal point for model training.

We further examined the relationship between computation time and MSE for both estimation approaches, as summarized in Table~\ref{time_compare}. In our experiments, the RNN (LSTM) produced parameter estimates in approximately $0.0072$ seconds, which is more than six times faster than the MOM, which required around $0.0429$ seconds. This comparison highlights that the RNN not only achieves better accuracy but also offers substantial computational efficiency during inference.

\begin{table}[H]
\centering
\begin{tabular}{|c|c|c|}
\hline
\textbf{Method} & \textbf{MSE} & \textbf{Time (sec)} \\
\hline
RNN (LSTM) & 0.6737 & 0.0072 \\
MOM        & 1.3033 & 0.0429 \\
\hline
\end{tabular}
\caption{Comparison of RNN (LSTM) and MOM in terms of mean squared error (MSE) and computation time.}
\label{time_compare}
\end{table}

Next, we investigate the statistical properties of the RNN (LSTM) estimator and examine its sampling distribution. For this purpose, 1000 paths were generated, each containing a sequence length of 30 events, using the true parameter values $\mu = 2.622$ and $\beta = 0.520$. The mean and standard deviation of the estimated parameters obtained from both RNN (LSTM) and MOM are summarized in Table~\ref{table_rnn_mom}. The results clearly show that the RNN (LSTM) estimator provides more accurate and stable parameter estimates, with lower variance compared to the method of moments, further confirming its robustness in modeling FPP dynamics.

\begin{table}[H]
\centering
\begin{tabular}{|c|c|cc|cc|}
\hline
\textbf{Parameter} & \textbf{True} & \multicolumn{2}{c|}{\textbf{RNN (LSTM)}} & \multicolumn{2}{c|}{\textbf{MOM}} \\ 
\hline
 &  & \textbf{Mean} & \textbf{S.D.} & \textbf{Mean} & \textbf{S.D.} \\ 
\hline
$\mu$   & 2.622 & 2.799 & 0.685 & 3.855 & 1.251 \\
$\beta$ & 0.520 & 0.502 & 0.084 & 0.347 & 0.086 \\
\hline
\end{tabular}
\caption{Parameter estimation summary for the RNN (LSTM) and method of moments (MOM).}
\label{table_rnn_mom}
\end{table}

 \section{Application on real data}\label{section:5}
 To validate the practical applicability of our method, we applied the trained LSTM 
to two real-world high-frequency datasets exhibiting temporal clustering and potential 
long-range dependence. To discuss the empirical results,  we consider two distinct datasets. The first dataset, MontcoAlert 911, contains real emergency call records from Montgomery County, PA (2015–2020), including the time, location, and type of incident. For the present analysis, we exclusively focused on the temporal aspect of the data. The timestamps were converted into inter-arrival times in seconds, representing the time gaps between successive emergency calls, and any zero or negative values were removed to ensure valid sequences. Using a sliding window of length 10, we estimate the FPP parameters $\mu$ and $\beta$ for each window via the MOM, considering these as the true values of the parameters. \\

 The second dataset corresponds to the NBBO quotes of AAPL, containing high-frequency intraday transaction timestamps for a specific trading day. The timestamps were converted into microseconds, and the differences between consecutive timestamps were used as inter-arrival times. These sequences were divided into overlapping sliding windows of length 50, and the FPP parameters $\mu$ and $\beta$ were estimated for each window using MOM. \\
 The LSTM was trained with the windowed inter-arrival sequences as input features and the MOM estimates as target variables. The RNN (LSTM) predictions closely follow the MOM values and produce smooth and consistent trajectories, showing that it can capture the intraday dynamics of high-frequency real datasets.

 \begin{figure}[H]
\centering
\begin{minipage}{.5\textwidth}
  \centering
  \includegraphics[width=1\linewidth]{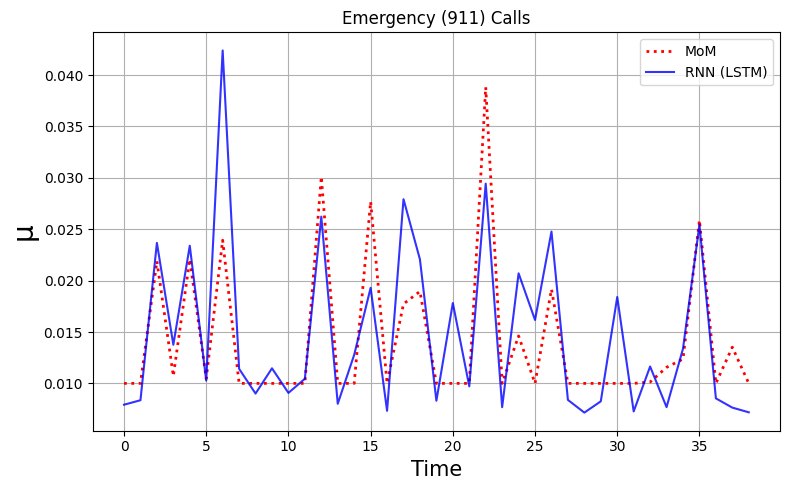}
  \label{mu_tfpp_911}
\end{minipage}%
\begin{minipage}{.5\textwidth}
  \centering
  \includegraphics[width=1\linewidth]{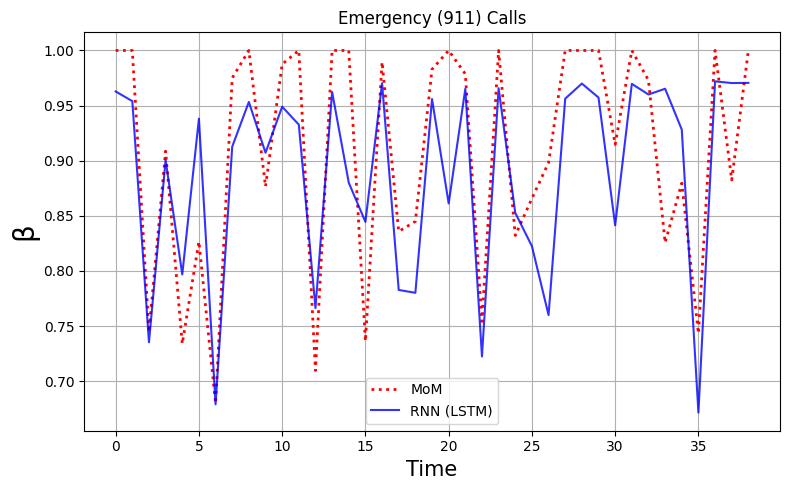}
  \label{beta_tfpp_911}
\end{minipage}
\caption{Time-varying parameters of the Emergency (911) calls dataset. (Top) The estimated intensity parameter 
$\mu >0$ over time using MOM (red dotted) and RNN (LSTM, blue solid). (Bottom) The estimated FPP
$\beta \in (0, 1)$ over time using MOM (red dotted) and RNN (LSTM, blue solid).}
 \end{figure}
\begin{figure}[H]
    \centering
    \includegraphics[width=1\linewidth]{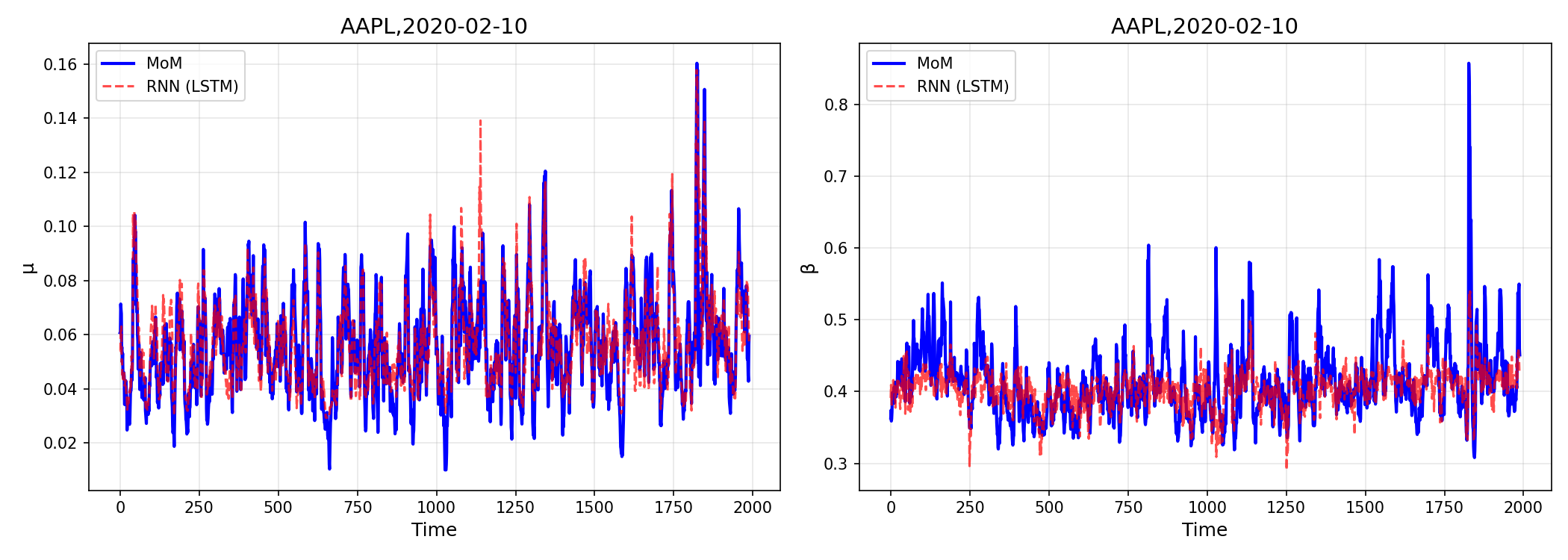}
    \label{mu_beta_aapl}
    \caption{Comparison of parameter estimates for AAPL $(2020-01-02)$. MOM (blue) is taken as the true reference, while the LSTM model (red) learns to approximate $\mu$ and $\beta$ from historical inter-arrival times.}
\end{figure}

\section*{Conclusion}\label{section:6}
The proposed LSTM-based RNN provides a robust and efficient solution for parameter estimation in complex fractional temporal point processes. It consistently outperforms the MOM in terms of MSE across various scenarios, demonstrating higher accuracy. Ablation studies confirm that the model is stable and reliable under different training conditions. Moreover, it is computationally efficient and performs effectively on real high-frequency datasets, capturing complex temporal patterns. Overall, these results highlight the LSTM-based approach as a practical, scalable, and reliable framework for parameter estimation, with strong applicability to real-world high-frequency data.
\section*{Acknowledgment}
We would like to thank Prof. Kyungsub Lee for kindly sharing the high-frequency finance data used in this research.

\def\cprime{$'$}

\end{document}